# Knowledge-Based Construction of Confusion Matrices for Multi-Label Classification Algorithms using Semantic Similarity Measures

Houcemeddine Turki [1], Mohamed Ali Hadj Taieb [1] and Mohamed Ben Aouicha [1]

[1] *Data Engineering and Semantics Research Unit, Faculty of Sciences of Sfax, University of Sfax, Sfax, Tunisia*

**Abstract**
So far, multi-label classification algorithms have been evaluated using statistical methods that do not consider the semantics of the considered classes and that fully depend on abstract computations such as Bayesian Reasoning. Currently, there are several attempts to develop ontology-based methods for a better assessment of supervised classification algorithms. In this research paper, we define a novel approach that aligns expected labels with predicted labels in multi-label classification using ontology-driven feature-based semantic similarity measures and we use it to develop a method for creating precise confusion matrices for a more effective evaluation of multi-label classification algorithms.

**Keywords**
Multi-Label Classification, Confusion Matrix, Evaluation, Semantic Similarity, Semantic Alignment, Accuracy Rate

## 1. Introduction

Supervised classification has currently become one of the most important challenges of machine learning with various applications in industry, medicine and other fields [1]. It consists of the controlled identification of given characteristics of an item such as a biomedical text or a clinical image [1]. There are two types of supervised classification: multi-label classification [2] and single-label multiclass classification [3]. Supervised Multi-label Classification assigns multiple labels in natural language to an item according to its characteristics while multiclass classification attributes one label in natural language to an item according to the measure of a given pattern [2, 3]. The evaluation of supervised classification algorithms primarily relies on the simple counts of *true positives* (TP), *true negatives* (TN), *false positives* (FP) and *false negatives* (FN) for each class [4, 5]. For a given class, the true positives are the cases where the class is accurately assigned to an item and the true negatives are the cases where the class is accurately unassigned to an item [4, 5]. These two counts combined with false positives (Unexpected classes predicted by the ML algorithm) and false negatives (Expected classes not returned by the ML algorithm) are integrated together to form three measures of the efficiency of classification algorithms: *precision*, *recall*, and *sensitivity* [4, 5]. These combined measures can be merged together to form curves and metrics for an overall evaluation of classification algorithms: *F1-measure*, *accuracy rate*, *Precision-Recall* (PR) curve and *Receiver Operating Characteristic* (ROC) curve [4, 5] as well as the Hamming loss[2] specific to the evaluation of multi-label learning [5, 6].

Beyond these classical approaches, the confusion matrix provides an interesting snapshot of the trends of a mono-label classification by computing the associations between true labels and predicted

---

---

[2] Hamming loss is the quotient of the number of non-alignments between true labels and predicted labels in multi-label classification out of the product of the number of considered classes with the number of classified items.

ones [4]. It identifies the set of classes that are not effectively differentiated and can provide several directions for explaining the limitations of assessed approaches for supervised learning and proposing solutions to them. Despite the easiness of the construction of a confusion matrix for mono-label classification [4], its creation for multi-label classification algorithms is tricky because of the difficulty of alignment between predicted classes and expected ones [7]. To solve this deficiency, a method has been developed based on Bayesian Reasoning to create fuzzy confusion matrices that approximately associate between true positives and false positives [7]. This approach is probabilistic and cannot assign with a full precision the false predicted classes corresponding to expected ones [7]. Here, the labels attributed by the multi-label classification algorithms are generally nouns or noun phrases having a semantic value [2, 3]. Consequently, semantic similarity measures (*SemSim*), particularly the ones based on taxonomies, can be useful to align true labels with predicted labels for a direct construction of precise confusion matrices for supervised multi-label classification given that these metrics return high values for semantically related terms [8].

In this research paper, we investigate this claim by applying semantic similarity measures to an ad-hoc fabricated output of a multi-label classification algorithm. We will begin by providing an overview of semantic similarity measures (Section 2). Then, we will describe our proposed approach for using these important metrics to construct confusion matrices for supervised multi-label classification (Section 3.1). After that, we will outline the methods that will be used for the assessment of our proposal (Section 3.2) and we will illustrate the outcomes of our preliminary experimental study and discuss them with reference to scholarly publications (Section 4). Finally, we will draw conclusions for this research paper and give future directions for developing this work (Section 5).

## 2. Semantic similarity measures

Measuring the degree of Semantic Similarity (*SemSim*) aims to quantify the likeness between linguistic items, including concepts and polysemous words. This has been a great challenge in the field of Natural Language Processing (NLP), and it is considered as a subfield of Artificial Intelligence focusing on the handling of human language by computers. As it is illustrated by Figure 1, computing the degree of semantic similarity between items faces two major challenges. First, it is necessary to build an appropriate word sense representation which has a fundamental impact on the efficiency of the estimation of semantic similarity, as a consequence of the expressiveness of the representation. The word sense representation is based on the extraction of information from semantic resources [9, 10] in several types: structured (WordNet, MeSH, Gene Ontology), semi-structured (Wikipedia, Wiktionary, etc.) and raw texts (corpora). The gathered information pertains to a large range of types such as the topological parameters, distributional semantic and word embedding [10, 11]. Then, the second challenge is the computing model aggregating between the different information according to their weights and semantic interpretation for providing the semantic similarity estimation. Selecting the appropriate semantic measure for improving the performance of an application depends on the nature of the application and the underlying knowledge source [10, 11]. The evaluation protocol of semantic similarity measures follows in totality or partially three approaches [12, 13]: *Intrinsic evaluation* based on datasets composed of a set of word pairs which their similarities are estimated by experts[3], *Semi-intrinsic evaluation approach* exploiting applications derived from word similarity such sentence and short text similarity tasks, and *Extrinsic assessment* involving the *SemSim* measure in specific applications like relation extraction, text summarization, sentiment analysis, word sense disambiguation, plagiarism detection, etc. An overview on the recent and previous works leads to the existence of two families of approaches as shown in Figure 1: the first is based on the structures of knowledge bases [11, 14] and content, and the second finds its root in the distributional semantics that have later contributed to the appearance of embedding methods [13, 15].

Knowledge-based measures exploit the topological parameters (depth[4], hyponyms[5], hypernyms[6] and lowest common subsumer[7]) of the semantic network modeled as direct acyclic graphs. Therefore,

---

[3] Accurate *SemSim* measures should generate closer similarities to those assigned by experts.

[4] The depth of a concept is the length of the longest path connecting the root of the taxonomy to the target concept. It assigns a value of 1 to the first-order metaclass (the common hypernym of all the concepts of the reference taxonomy) and a value of $N$ to the $N^{th}$-order

several measures judged as structural approaches exploited the taxonomic parameters extracted from the "is a" taxonomy. Several measures for determining the semantic similarity between words/concepts have been proposed in the literature and most of them have been tested using WordNet. Similarity measures are based on the overlapped features designed through the "is a" taxonomy. The measures can be grouped into five classes: path-based measures, gloss-based measures, feature-based measures, information content (IC)-based measures and hybrid measures [11, 17]. Path-based measures only compute the shortest path between two terms in an "is a" taxonomy, sometimes in function of the depth of concepts in the reference resource [17]. Consequently, they are unable to distinguish homologous terms where the link between them is non-taxonomic (e.g., Drugs treating similar diseases). Gloss-based measures compare the named entities included in the glosses (i.e., definitions) of two terms to identify if these two concepts are semantically related [11, 18]. As a result, these metrics can return limited results if the glosses of compared entities in a reference thesaurus are brief or unavailable [11, 18]. IC-based measures calculate the semantic similarity of items according to their co-occurrence in a large textual corpus [17]. Further that running this type of metrics is more time-consuming than the ones based on taxonomies or thesauri, this kind of measures is probabilistic and can return different results according to the used corpus for reference [17]. Feature-Based Measures compare two terms according to a various set of characteristics in the reference knowledge resource including their taxonomic and non-taxonomic relationships to other terms and the similarity between the concepts included in their glosses [17, 19]. Thanks to their principle, these metrics are the best ones that can be used in our context as they return homologous concepts according to various types of semantic links [17, 19]. As these metrics perform the best in our situation, coupling them with other types of semantic similarity measures to form hybrid measures can negatively influence their efficiency to return homology between two concepts [17]. Given the characteristics of every type of semantic similarity measures, feature-based metrics are the ones that should be used to identify homologous terms and consequently to align expected labels with predicted ones in multi-label classification.

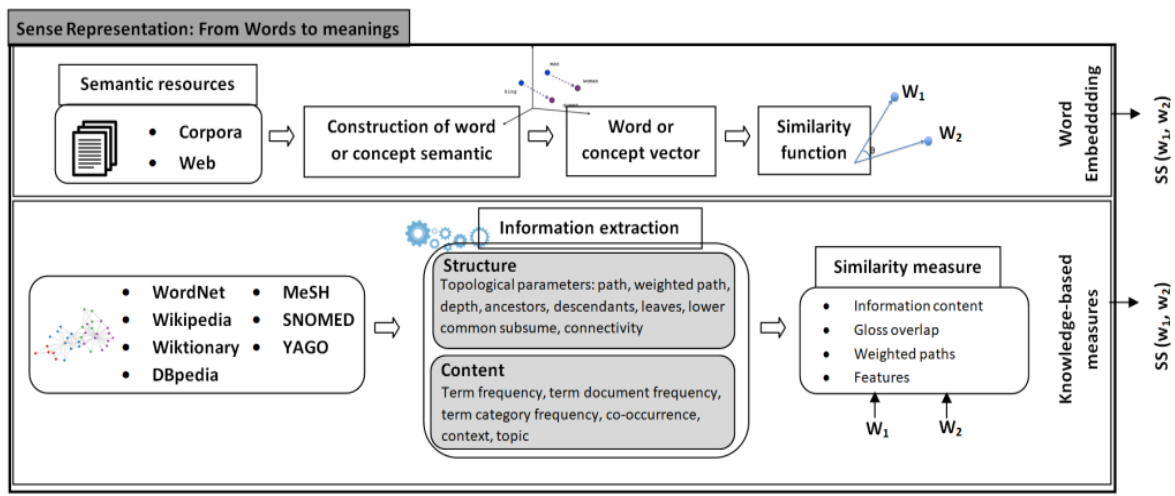

**Figure 1**: Process exploiting knowledge-based and/or distributional approaches for computing semantic similarity between words/concepts

## 3. Proposed Approach

In this section, we introduce the principles of our approach that uses semantic similarity measures to assign predicted labels to corresponding expected labels and consequently to allow the easy construction of a confusion matrix for supervised multi-label classification (Section 3.1). Then, we

---

concept (i.e. a concept that is linked to the first-order metaclass through a hierarchy of (N-1) hypernyms). To compute the taxonomic depth of a concept in WordNet, you can use the WNetSS API available at https://github.com/MohamedAliHadjTaieb/WNetSS-API/ [16].
  [5] Hyponym: a direct or an indirect descendant of a given term.
  [6] Hypernym: a direct or an indirect parent of a given term.
  [7] Lower Common Subsumer: the closest common parent of two given terms.

explain the experimental methods that will be applied for testing our approach on four practical examples that imitate the real-world outputs of supervised multi-label classification algorithms (Section 3.2).

## 3.1. Principles

In our approach, we propose to compute the semantic similarity between each predicted label ($x = p_i$) and each expected one ($e_i$) for every assessed item. Then, we will align between predicted and expected labels using a set of four rules where $N_P$ is the number of predicted labels, $N_E$ is the number of expected labels for a given item, $F_N$ is the value returned by *SemSim* for two absolutely non-similar terms (minimal value of *SemSim*), and $F_M$ is the value returned by *SemSim* for two synonym terms (maximal value of *SemSim*).

**Rule 1:** If $N_P > N_E$, for every true label ($Y = e_i$), we assign the predicted label ($P$) that returns the highest value of semantic similarity between $Y$ and $x$ as the corresponding one to the expected label ($Y$) [Equation 1]:

$$P = \arg\max(SemSim(x, Y)) \quad (1)$$

**Rule 2:** If $N_P \leq N_E$, for every predicted label ($P = p_i$), we assign the expected label ($Y$) that returns the highest value of semantic similarity between $P$ and $x$ as the corresponding one to the predicted label ($P$) [Equation 2]:

$$Y = \arg\max(SemSim(x, P)) \quad (2)$$

**Rule 3:** To prevent the assignment of an association between unrelated labels, we do not consider the link between an expected class (Y) and a predicted class (P) when $SemSim(P, Y) < 0.5 * (F_M + F_N)$. As distinct *SemSim* measures can have distinct minimal and maximal values [11, 14], we cannot set a real number as a constant threshold for all semantic similarity measures.

**Rule 4:** When two associations involve the same expected label or the same predicted one, only the association with the highest value of *SemSim* is kept. The eliminated association should be substituted by the second sorted association according to the rules used for its recognition (Rules 1, 2 and 3).

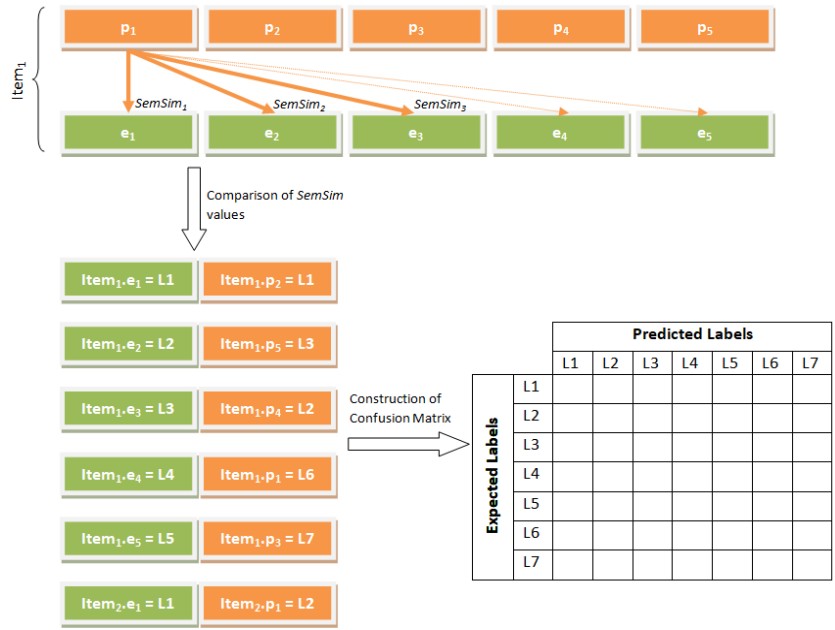

**Fig. 2.** Process of the proposed semantic similarity-based approach

Consequently, we obtain a list of correspondence between predicted labels (*P*) and expected labels (*Y*) in supervised multi-label classification that can be used to create a confusion matrix for multi-label classification similarly as the one for mono-label classification as shown in Fig. 2. Values in confusion matrices can be either simple counts of the associations between expected labels and predicted ones or the rate of expected labels associated with each predicted one [4]. In this study, we will use simple counts of the coupled labels to construct the confusion matrix based on the semantic similarity-based alignment of expected classes with predicted ones. If this matrix is properly constructed, it can be useful to identify the set of labels that are not efficiently distinguished by multi-label classification algorithms and add interesting information to the explanation of the accuracy rates for such algorithms [4].

## 3.2. Experimental Study

Through a series of practical applications, we try to validate the assumption that the application of semantic similarity measures to compare expected labels and predicted ones for each analyzed item can serve to align predicted labels with corresponding true ones and consequently to develop a confusion matrix for supervised multi-label classification. Despite the significance of this matter, there is no human-generated dataset that aligns between false predicted labels and expected labels in the output of a multi-label classification. Although such a dataset can be created using an annotated image database like Tencent ML Images [20], this process cannot be practically done as this requires lots of resources and efforts. In fact, the dataset that we need for our analysis should:

1. Provide expected classes for every item and this is available for various online databases like Tencent ML Images.
2. Return the classes that have been predicted by a multi-label classification algorithm for every item and this is possible by applying machine learning models on multi-label classification datasets.
3. Map the considered classes to their corresponding concepts in a reference knowledge database, particularly an "is a" taxonomy so that the computation of semantic similarity measures between classes can be enabled. This can be easy for datasets using controlled vocabularies to annotate items. However, it will be difficult to perform such a semantic mapping for classes generated by dataset creators and not inferred from a common semantic resource.
4. Manually align between false positives and corresponding false negatives. This will need a lot of human contributors to process tens of considered classes for thousands of classified items. As well, this will need a lot of field expertise so that classes that represent the same attribute can be easily recognized.

Instead, we build an ad-hoc dataset of items that are assigned expected labels as well as sets of predicted labels that look like as if they were attributed by a multi-label classification algorithm as shown in Table 1. There is no application of supervised multi-label classification algorithms in this study and all the dataset is fabricated for the evaluation of our approach. The labels that are assigned to the virtual items have been derived from the WordNet 3.1 taxonomy [21], specifically from the lexicon related to animals. As shown in Table 1, the terms are focused on canids, particularly dogs and wolfs, and on the direct hyponyms of the "animal" term.

**Table 1**
Expected and Predicted Labels for virtual items made using WordNet 3.1

| Item | Expected Labels | Predicted Labels |
|---|---|---|
| 1 | Animal, Canid, Dog, Domesticated animal, German Shepherd[A1], Pet | Animal, Belgian Shepherd[A1], Canid, Dog, Domesticated animal, Pet |
| 2 | Animal, Canid, Dog[A2], Domesticated animal, German Shepherd[A1], Pet | Animal, Belgian Shepherd[A1], Canid, Domesticated animal, Pet, Wolf[A2] |
| 3 | Animal, Canid, Dog[A2], Domesticated | Animal, Belgian Shepherd[A1], Canid, Pet, |
| 4 | animal[(A3)], German Shepherd[(A1)], Pet Animal, Canid, Dog[(A2)], Domesticated animal, German Shepherd[(A1)], Pet[(A4)] | Wolf[(A2)], Work animal[(A3)] Animal, Belgian Shepherd[(A1)], Predatory animal[(A4)], Wolf[(A2)] |

The constructed examples consider the complexity of the matter of aligning true labels with predicted labels in multi-label classification. Effectively, the first example (Item 1) that just substitutes one label by another one (A1) is considered as the simplest one. By contrast, the fourth example (Item 4) that changes three labels by other closely related ones (A1, A2 and A4) and eliminates two classes is considered the most complicated example as it does not only assess the capacity of semantic similarity measures to align labels but also the robustness of our approach to the omission of labels by the multi-label classification algorithm. The second (Item 2) and third (Item 3) examples respectively substitute two labels (A1 and A2) and three labels (A1, A2 and A3) by other significantly linked classes and are considered as moderately difficult samples of the input of the systems for the alignment between predicted and true labels in supervised multi-label classification.

To test our assumption, we will use the feature-based measure described in Rodriguez et al. (2003)[8] to compute the similarity between predicted classes and expected ones [19]. This method returns a value ranging between 0 (= $F_N$) and 3 (= $F_M$) [19] and is designed to identify similar items according to the specificity of their semantic characteristics rather than their inclusion to the same lexical field in the context of topic modeling and data mining [11]. Further theoretical details about this measure can be found in Hadj Taieb et al. (2014) [14]. We drive this similarity metric by the *WordNet 3.1* taxonomy that is also used to construct the examples. The annotation of items with classes derived from structured knowledge resources, particularly controlled taxonomies, and the use of the same lexical database to drive the adopted knowledge-based semantic similarity measure as well as to annotate the considered items for supervised multi-label classification will allow getting precise evaluation of semantic similarity by preventing confusion between the assessed terms and the concepts in the reference resource. After getting the values of semantic similarity measures between predicted labels and true ones for the four examples, we will align between expected labels and predicted ones and finally construct the confusion matrix for these fabricated examples as explained in the Section 3.1.

## 4. Results and Discussion

The application of the metric of Rodriguez et al. (2003) to compare the expected labels and predicted ones that are assigned to item 1 (Table 2) and item 2 (Table 3) has shown that the efficiency of feature-based semantic similarity measures to align true labels and predicted ones with an absolute accuracy. Although the linking between expected and predicted classes for the item 1 is slightly challenging due to the existence of a unique difference between the two sets of labels, the precision of Rules 1 and 2 to attribute two false positives to their corresponding false negatives (bold in Table 3) proves the promising value of using semantic similarity measures to exhaustively relate between expected and predicted labels in multi-class classification, particularly because all the considered labels are linked to animals and are closely related together from a semantic perspective (Table 1). Yet, this does not prove that all the semantic similarity measures can be used for such a purpose. The computation of semantic similarity between too closely related terms has always been a challenge to the semantic web community [22] and not all the semantic similarity measures behave the same when applied to a given couple of terms [11, 14]. The success of Rodriguez et al. (2003) in such a task proves the added value of feature-based metrics among other semantic similarity measures in discriminating two quite similar terms [22]. Further investigation in this context can expand our findings.

In another circumstance, the outputs of tables 2 and 3 also show that the labels that are meant to be aligned together have values of semantic similarity that are superior or equal to 1.5. Most of the labels that are unrelated are assigned low values of semantic similarity mostly below 1. This motivates the use of Rule 3 to eliminate the associations that are weak and not accurate. It is true that most of the applications of semantic similarity measures, particularly in information retrieval and word sense disambiguation, use a threshold that is higher than $0.5 * (F_M + F_N)$ [23]. However, this threshold seems to work well for our approach especially for aligning "dog" as an expected label with "wolf" as a predicted one for item 2 where the value of semantic similarity is 1.789 as shown in Table 3.

---
[8] We will use https://github.com/MohamedAliHadjTaieb/WNetSS-API/ as a tool to compute the values of Semantic Similarity as returned by Rodriguez et al. (2003) [16].

**Table 2**
Semantic similarity measures between predicted and expected labels for item 1[9]

|  |  | Predicted Labels |  |  |  |  |  |
|---|---|---|---|---|---|---|---|
|  |  | Animal | Belgian Shepherd | Canid | Dog | Domesticated animal | Pet |
| True Labels | Animal | **3.000** | 0.417 | 0.707 | 0.748 | 1.502 | 1.513 |
|  | Canid | 0.792 | 1.032 | **3.000** | 1.569 | 1.057 | 0.833 |
|  | Dog | 0.737 | 1.085 | 1.576 | **3.000** | 0.977 | 1.330 |
|  | Domesticated animal | 1.471 | 0.482 | 0.996 | 0.883 | **3.000** | 1.872 |
|  | German Shepherd | 0.643 | **2.000** | 1.025 | 1.070 | 0.644 | 0.625 |
|  | Pet | 1.474 | 0.441 | 0.720 | 1.354 | 1.872 | **3.000** |

**Table 3**
Semantic similarity measures between predicted and expected labels for item 2

|  |  | Predicted Labels |  |  |  |  |  |
|---|---|---|---|---|---|---|---|
|  |  | Animal | Belgian Shepherd | Canid | Domesticated animal | Pet | Wolf |
| True Labels | Animal | **3.000** | 0.417 | 0.707 | 1.502 | 1.513 | 0.977 |
|  | Canid | 0.792 | 1.032 | **3.000** | 1.057 | 0.833 | 1.574 |
|  | Dog | 0.737 | 1.085 | 1.576 | 0.977 | 1.330 | **1.789** |
|  | Domesticated animal | 1.471 | 0.482 | 0.996 | **3.000** | 1.872 | 0.783 |
|  | German Shepherd | 0.643 | **2.000** | 1.025 | 0.644 | 0.625 | 1.023 |
|  | Pet | 1.474 | 0.441 | 0.720 | 1.872 | **3.000** | 1.020 |

The use of our rule-based approach to match the expected and predicted labels of the item 3 confirms our findings about the usefulness of feature-based semantic similarity measures for linking between the true classes and the predicted ones in supervised multi-label classification and the efficiency of our three first rules for performing this mission. Effectively, as shown in Table 4, most of the labels that are confused together are assigned high values of semantic similarity. The association between "Work animal" and "Domesticated animal" has been identified instead of the one between "Work animal" and "Pet" despite the latter has a higher semantic similarity because of the application of Rule 4. This condition considered that "Pet" as an expected label has already been matched to "Pet" as a predicted one and consequently disregarded any other possible association between "Pet" and other classes. Consequently, this rule is important to avoid mismatches between predicted and expected labels, mainly the ones due to the fact that semantic similarity measures assign high values of semantic similarity between a concept and its hypernym [14]. Without this rule, "Pet" as an expected label would be assigned to "Pet" and "work animal" as predicted ones. This would not let users precisely study the confusion between classes in a reliable way and the usage of semantic similarity measures for our work will be quite inefficient.

---

[9] Bold values correspond to the identified associations between predicted labels and expected ones based on the four rules defined in the Section 3.1.

Table 4

Semantic similarity measures between predicted and expected labels for item 3

|  |  | Predicted Labels | | | | | |
|---|---|---|---|---|---|---|---|
|  |  | Animal | Belgian Shepherd | Canid | Pet | Wolf | Work animal |
| True Labels | Animal | **3.000** | 0.417 | 0.707 | 1.513 | 0.977 | 1.513 |
|  | Canid | 0.792 | 1.032 | **3.000** | 0.833 | 1.574 | 0.833 |
|  | Dog | 0.737 | 1.085 | 1.576 | 1.330 | **1.789** | 0.888 |
|  | Domesticated animal | 1.471 | 0.482 | 0.996 | 1.872 | 0.783 | **1.872** |
|  | German Shepherd | 0.643 | **2.000** | 1.025 | 0.625 | 1.023 | 0.625 |
|  | Pet | 1.474 | 0.441 | 0.720 | **3.000** | 1.020 | 1.999 |

The efficiency of our approach has been confirmed when used to identify the associations between true and predicted classes for item 4. Despite three class substitutions and two omitted labels for this item by the conceived multi-label classification, the semantic similarity measure has been successful to fully identify all the associations between predicted and expected labels (Bold in Table 5). It is true that Rule 3 limits the recognition of valid links between false positives and false negatives where these statistical co-occurrences are not coupled to a semantic similarity between the confused concepts (e.g., a horse recognized as a potato in a blurry image). However, this rule was behind the elimination of the noisy association between unrelated concepts and consequently the precise identification of associations between false positives and false negatives. This proves the robustness of our method to the non-annotation of accurate labels by humans or to the exclusion of several classes by the supervised classification algorithms. This can be a significant contribution to the longstanding challenge of developing a large-scale supervised multi-label learning system that can deal with missing labels [24] as well as to the differentiation between substituted classes and missing ones in supervised multi-label classification. The extent of importance of semantic similarity measures to the development of supervised multi-label classification systems considering missing labels depends on the assessment of our approach on more complex examples than the four that have been developed for this research paper (Table 1). Unfortunately, there is still a significant lack of databases providing outputs of supervised multi-label classification systems and originally assigned labels for each item at a large scale to allow the assessment of the behavior of our method to face all the kinds of differences between predicted and expected labels. It is evident that such a database can be created by applying a classification algorithm to a multi-labeled dataset [25]. However, this will bring challenges to develop thorough semantic similarity-driven approaches for supervised multi-label classification learning and evaluation.

Table 5

Semantic similarity measures between predicted and expected labels for item 4

|  |  | Predicted Labels | | | |
|---|---|---|---|---|---|
|  |  | Animal | Belgian Shepherd | Predatory animal | Wolf |
| True Labels | Animal | **3.000** | 0.417 | 1.513 | 0.977 |
|  | Canid | 0.792 | 1.032 | 0.833 | 1.574 |
|  | Dog | 0.737 | 1.085 | 0.889 | **1.789** |
|  | Domesticated animal | 1.471 | 0.482 | 1.872 | 0.783 |
|  | German Shepherd | 0.643 | **2.000** | 0.625 | 1.023 |

| | Pet | 1.474 | 0.441 | **2.000** | 1.020 |
|---|---|---|---|---|---|

Due to the ability of semantic similarity measures to distinguish between related and unrelated classes in supervised multi-label classification, the outputs of the tables 2 to 5 can be easily processed allowing the construction of a confusion matrix for the four considered examples (Items 1 to 4) from the identified alignments between predicted and expected labels (Bold in Tables 2 to 5) as shown in Table 6. It is accurate that a method for generating a fuzzy confusion matrix for supervised multi-label classification algorithms has already been developed based on the pairwise label transformation [7]. This method turns the supervised multi-label classification problem into a set of binary classifications, assesses the efficiency of the classifier for performing each binary classification, and then traces a confusion matrix that returns the accuracy rate of the classification algorithm to distinguish between each pair of assigned labels [7]. However, our semantic similarity-driven approach for creating a confusion matrix for multi-label classification seems to be simpler and computationally easier than the fuzzy confusion matrix method as it easily identifies the labels that have been correctly annotated (Grey in Table 6) and points out the labels that are confused together with a high accuracy (Bold and not grey in Table 6). The computation of semantic similarity between two terms requires less time than running a binary classification on a full dataset [26]. That is why it can be a good alternative for promoting research on semantics-aware assessment of supervised multi-label classification systems rather than continuing in developing methods for the construction of confusion matrices based on fully statistical and imprecise approaches. Effectively, many works are recently developed following the path of using semantics for the development and evaluation of machine learning algorithms as shown in Table 7.

**Table 6**
Confusion matrix based on the semantic similarity-based associations between the predicted labels and expected ones in the four examples

| | | Expected Labels | | | | | |
|---|---|---|---|---|---|---|---|
| | | Animal | Canid | Dog | Domesticated animal | German Shepherd | Pet |
| Predicted Labels | Animal | **4** | 0 | 0 | 0 | 0 | 0 |
| | Belgian Shepherd | 0 | 0 | 0 | 0 | **4** | 0 |
| | Canid | 0 | **3** | 0 | 0 | 0 | 0 |
| | Dog | 0 | 0 | **1** | 0 | 0 | 0 |
| | Domesticated animal | 0 | 0 | 0 | **2** | 0 | 0 |
| | German Shepherd | 0 | 0 | 0 | 0 | 0 | 0 |
| | Pet | 0 | 0 | 0 | 0 | 0 | **3** |
| | Predatory animal | 0 | 0 | 0 | 0 | 0 | **1** |
| | Wolf | 0 | 0 | **3** | 0 | 0 | 0 |
| | Work animal | 0 | 0 | 0 | **1** | 0 | 0 |
| | *No value* | 0 | 1 | 0 | 1 | 0 | 0 |

**Table 7**
Existing approaches for Semantics-driven evaluation of supervised classification

| Method | Description | Example |
|---|---|---|

| *Propensity-scored losses* [6] | Adjusted editions of Hamming loss that do not compute the non-correspondence between an estimated class and the true classes of an item if a true class is a subclass of the estimated class according to a given taxonomy such as the Wikipedia Category Graph. This reduces the effect of missing labels in the reference dataset allowing a more consistent accuracy evaluation for the multi-label supervised classification algorithms. | **Expected Labels:** *Cat* and *Persian cat*<br>**Predicted Labels:** *Persian cat*<br>⇨ *Persian cat* is a subclass of *Cat*. Therefore, *Cat* is considered as if it was predicted by the algorithm |
|---|---|---|
| *Hierarchical Multi-Label Classification* [27] | A statistical approach to split a supervised multi-label classification problem into a set of multiple mono-label classification problems according to the analysis of label associations in the training dataset. This method narrows the complexity of the original multi-label classification and debugging the obtained mono-label classifications one-by-one through their assessment with confusion matrices. | **Considered Classes:** *Yellow*, *White*, *Black*, Grey, *Persian Cat*, *Bengal Cat*, *Munchkin Cat*, and *Scottish Fold*<br>**Generated Classifications:**<br>• First Classification (Cat Colors): *Yellow*, *White*, *Black*, and *Grey*.<br>• Second Classification (Cat Breeds): *Persian Cat*, *Bengal Cat*, *Munchkin Cat*, and *Scottish Fold*. |
| *Inductive Multi-Label Classification* [28] | A semantic approach that uses "part of" relations to deduce the existence of an item where its components are identified. Such a process allows to recover the concepts that were missed by the algorithm where their features are recognized. | *Eye*, *Nose* and *Mouth* are recognized in a given image.<br>⇨ {*Eye*, *Nose*, *Mouth*} are parts of *Face*. That is why *Face* should be recognized in the assessed image. |
| *Constraint-Based Evaluation of Semantic Relations* [29] | An approach that evaluates if the learned type of a semantic relation is conflicting with logical constraints in a reference knowledge graph. This method allows the validation of semantic relations from a structural and an axiomatic perspective thanks to a set of formally defined conditions (TBox) coupled to statistically asserted conditions (ABox). | **Learned Relation:** *Hepatitis C* is a *drug used for treatment* of *Sofosbuvir*<br>**Constraints:**<br>• Subject should be instance of *drug*<br>• Object should be instance of *disease*<br>**Evaluation:** False (Subject-Object Inversion) |

## 5. Conclusion

In this research paper, we presented feature-based semantic similarity measures as useful components for the construction of confusion matrices for multi-label classification learning algorithms. Then, we demonstrated a promising efficiency of our proposed approach by applying it on four practical examples that have been preliminarily created based on the WordNet 3.1 taxonomy. This work is a development of the sustainable efforts to develop more effective approaches for the explanation of the limitations of multi-label classification algorithms [4-6, 27]. That is why we invite scientists to develop our approach as it can be efficient in enhancing works on artificial intelligence. As a future direction of this research work, we will develop other applications of semantic similarity measures and graph embeddings to solve other critical matters in Artificial Intelligence. Effectively, graph embeddings allow the construction of a consistent database for feature representations (feature vectors and spaces) that can be later applied on datasets for classification purposes [30]. The analysis of such inferred features can be very useful to recognize several fallacies in the considered ML model [31]. We will also try to create a large-scale benchmark that can be adapted to assess multi-label classification evaluation methods and use it to reproduce our work using featured-based semantic similarity measures other than the one of Rodriguez et al. (2003) [19]. Such an analysis will allow to evaluate the consistency of semantic similarity measures in constructing confusion matrices for multi-label classification through the comparison of its outputs based on a large dataset to the ones of statistical approaches such as pairwise confusion matrix construction and Bayesian confusion matrix creation.

## 6. References

[1] J. Du, Q. Chen, Y. Peng, Y. Xiang, C. Tao, Z. Lu, ML-Net: multi-label classification of biomedical texts with deep neural networks, Journal of the American Medical Informatics Association 26 (2019): 1279–1285. doi:10.1093/jamia/ocz085.

[2] L. Sun, H. Ge, W. Kang, Non-negative matrix factorization based modeling and training algorithm for multi-label learning, Frontiers of Computer Science 13 (2019): 1243–1254. doi:10.1007/s11704-018-7452-y.

[3] W. La Cava, S. Silva, K. Danai, L. Spector, L. Vanneschi, J. H. Moore, Multidimensional genetic programming for multiclass classification, Swarm and Evolutionary Computation 44 (2019): 260–272. doi:10.1016/j.swevo.2018.03.015.

[4] A. Tharwat, Classification assessment methods, Applied Computing and Informatics (2020). doi:10.1016/j.aci.2018.08.003.

[5] G. Tsoumakas, I., Katakis, Multi-Label Classification: An overview, International Journal of Data Warehousing and Mining 3 (2007): 1–13. doi:10.4018/jdwm.2007070101.

[6] H. Jain, Y. Prabhu, M. Varma, Extreme Multi-label Loss Functions for Recommendation, Tagging, Ranking & Other Missing Label Applications, in: Proceedings of the 22nd ACM SIGKDD International Conference on Knowledge Discovery and Data Mining, ACM, San Francisco, 2016, pp. 935–944. doi:10.1145/2939672.2939756.

[7] P. Trajdos, M. Kurzynski, Weighting scheme for a pairwise multi-label classifier based on the fuzzy confusion matrix, Pattern Recognition Letters 103 (2018): 60–67. doi:10.1016/j.patrec.2018.01.012.

[8] M. A. Hadj Taieb, M. Ben Aouicha, A. Ben Hamadou, A new semantic relatedness measurement using WordNet features, Knowledge and Information Systems 41 (2014): 467–497. doi:10.1007/s10115-013-0672-4.

[9] M. Ben Aouicha, M. A. Hadj Taieb, A. Ben Hamadou, Taxonomy-based information content and wordnet-wiktionary-wikipedia glosses for semantic relatedness, Applied Intelligence 45 (2016): 475–511. doi:10.1007/s10489-015-0755-x.

[10] R. Qu, Y. Fang, W. Bai, Y. Jiang, Computing semantic similarity based on novel models of semantic representation using Wikipedia, Information Processing & Management 54.6 (2018): 1002-1021. doi:10.1016/j.ipm.2018.07.002.

[11] J. J. Lastra-Díaz, J. Goikoetxea, M. A. Hadj Taieb, A. García-Serrano, M. Ben Aouicha, E. Agirre, A reproducible survey on word embeddings and ontology-based methods for word similarity: Linear combinations outperform the state of the art, Engineering Applications of Artificial Intelligence 85 (2019): 645–665. doi:10.1016/j.engappai.2019.07.010.

[12] M. A. Hadj Taieb, T. Zesch, M. Ben Aouicha, A survey of semantic relatedness evaluation datasets and procedures, Artificial Intelligence Review 53 (2020): 4407–4448. doi:10.1007/s10462-019-09796-3.

[13] M. Artetxe, G. Labaka, I. Lopez-Gazpio, E. Agirre, Uncovering Divergent Linguistic Information in Word Embeddings with Lessons for Intrinsic and Extrinsic Evaluation, in: Proceedings of the 22nd Conference on Computational Natural Language Learning, ACL, Brussels, Belgium, 2018, pp. 282-291. doi:10.18653/v1/K18-1028.

[14] M. A. Hadj Taieb, M. Ben Aouicha, A. Ben Hamadou, Ontology-based approach for measuring semantic similarity, Engineering Applications of Artificial Intelligence 36 (2014): 238–261 (2014). doi:10.1016/j.engappai.2014.07.015.

[15] J. Camacho-Collados, M. T. Pilehvar, From Word To Sense Embeddings: A Survey on Vector Representations of Meaning, Journal of Artificial Intelligence Research 63 (2018): 743–788. doi:10.1613/jair.1.11259.

[16] M. Ben Aouicha, M. A. Hadj Taieb, A. Ben Hamadou, SISR: System for integrating semantic relatedness and similarity measures, Soft Computing 22 (2016): 1855–1879. doi:10.1007/s00500-016-2438-x.

[17] T. Slimani, Description and Evaluation of Semantic Similarity Measures Approaches, International Journal of Computer Applications 80.10 (2013): 25-33. doi:10.5120/13897-1851.

[18] S. Banerjee, T. Pedersen, An adapted Lesk algorithm for word sense disambiguation using WordNet, in: International conference on intelligent text processing and computational linguistics, Springer, Berlin, Heidelberg, 2002, pp. 136-145. doi:10.1007/3-540-45715-1_11.

[19] M. A. Rodriguez, M. J. Egenhofer, Determining semantic similarity among entity classes from different ontologies, IEEE Transactions on Knowledge and Data Engineering 15 (2003): 442–456. doi:10.1109/tkde.2003.1185844.

[20] B. Wu, W. Chen, Y. Fan, Y. Zhang, J. Hou, J. Liu, T. Zhang, Tencent ml-images: A large-scale multi-label image database for visual representation learning, IEEE Access, 7 (2019): 172683-172693. doi:10.1109/ACCESS.2019.2956775.

[21] G. A. Miller, WordNet: An electronic lexical database, MIT Press, Cambridge, MA, 1998. ISBN:978-0-262-06197-1.

[22] A. Budanitsky, G. Hirst, Evaluating WordNet-based Measures of Lexical Semantic Relatedness. Computational Linguistics 32 (2006): 13–47. doi:10.1162/coli.2006.32.1.13.

[23] A. Hliaoutakis, G. Varelas, E. Voutsakis, E. G. M. Petrakis, E. Milios, Information Retrieval by Semantic Similarity, International Journal on Semantic Web and Information Systems 2 (2006): 55–73. doi:10.4018/jswis.2006070104.

[24] H. F. Yu, P. Jain, P. Kar, I. Dhillon, Large-scale multi-label learning with missing labels, in: Proceedings of the 31st International Conference on Machine Learning, PMLR, Beijing, 2014, pp. 593-601. doi:10.5555/3044805.3044873.

[25] A. Bustos, A. Pertusa, J.-M. Salinas, M. de la Iglesia-Vayá, PadChest: A large chest x-ray image dataset with multi-label annotated reports, Medical Image Analysis 66 (2020): 101797. doi:10.1016/j.media.2020.101797.

[26] J. J. Lastra-Díaz, A. García-Serrano, M. Batet, M. Fernández, F. Chirigati, HESML: A scalable ontology-based semantic similarity measures library with a set of reproducible experiments and a replication dataset, Information Systems 66 (2017): 97–118. doi:10.1016/j.is.2017.02.002.

[27] J. Wehrmann, R. Cerri, R. Barros, Hierarchical Multi-Label Classification Networks, in: Proceedings of the 35th International Conference on Machine Learning, PMLR, Stockholm, 2018, pp. 5075-5084.

[28] M. K. Sarker, J. Schwartz, P. Hitzler, L. Zhou, S. Nadella, B. Minnery, I. Juvina, M. L. Raymer, W. R. Aue, Wikipedia knowledge graph for explainable ai, in: Iberoamerican Knowledge Graphs and Semantic Web Conference, Springer, Cham., 2020, pp. 72-87. doi:10.1007/978-3-030-65384-2_6.

[29] M. Alshahrani, M. A. Thafar, M. Essack, Application and evaluation of knowledge graph embeddings in biomedical data, PeerJ Computer Science 7 (2021): e341. doi:10.7717/peerj-cs.341.

[30] A. Amara, M. A. Hadj Taieb, M. Ben Aouicha, Network representation learning systematic review: Ancestors and current development state, Machine Learning with Applications (2021). doi:10.1016/j.mlwa.2021.100130.

[31] A. Holzinger, B. Malle, A. Saranti, B. Pfeifer, Towards multi-modal causability with Graph Neural Networks enabling information fusion for explainable AI, Information Fusion 71 (2021): 28-37. doi:10.1016/j.inffus.2021.01.008.